# A Multiagent Reinforcement Learning Algorithm with Non-linear Dynamics

**Sherief Abdallah**                                  SHERIEF.ABDALLAH@BUID.AC.AE
*Faculty of Informatics*
*The British University in Dubai*
*United Arab Emirates*
*(Fellow) School of Informatics*
*University of Edinburgh*
*United Kingdom*

**Victor Lesser**                                     LESSER@CS.UMASS.EDU
*Department of Computer Science*
*University of Massachusetts Amherst*
*United States*

## Abstract

Several multiagent reinforcement learning (MARL) algorithms have been proposed to optimize agents' decisions. Due to the complexity of the problem, the majority of the previously developed MARL algorithms assumed agents either had some knowledge of the underlying game (such as Nash equilibria) and/or observed other agents actions and the rewards they received.

We introduce a new MARL algorithm called the Weighted Policy Learner (WPL), which allows agents to reach a Nash Equilibrium (NE) in benchmark 2-player-2-action games with minimum knowledge. Using WPL, the only feedback an agent needs is its own local reward (the agent does not observe other agents actions or rewards). Furthermore, WPL does not assume that agents know the underlying game or the corresponding Nash Equilibrium a priori. We experimentally show that our algorithm converges in benchmark two-player-two-action games. We also show that our algorithm converges in the challenging Shapley's game where previous MARL algorithms failed to converge without knowing the underlying game or the NE. Furthermore, we show that WPL outperforms the state-of-the-art algorithms in a more realistic setting of 100 agents interacting and learning concurrently.

An important aspect of understanding the behavior of a MARL algorithm is analyzing the dynamics of the algorithm: how the policies of multiple learning agents evolve over time as agents interact with one another. Such an analysis not only verifies whether agents using a given MARL algorithm will eventually converge, but also reveals the behavior of the MARL algorithm prior to convergence. We analyze our algorithm in two-player-two-action games and show that symbolically proving WPL's convergence is difficult, because of the non-linear nature of WPL's dynamics, unlike previous MARL algorithms that had either linear or piece-wise-linear dynamics. Instead, we numerically solve WPL's dynamics differential equations and compare the solution to the dynamics of previous MARL algorithms.

## 1. Introduction

The decision problem of an agent can be viewed as selecting a particular action at a given state. A well-known simple example of single-agent decision problem is the multi-armed bandit (MAB) problem: an agent needs to choose a lever among a set of available levers. The reward of executing each action is drawn randomly according to a fixed distribution. The agent's goal is to choose the lever (the action) with the highest expected reward. In order to do so, the agent samples the underlying reward distribution of each action (by trying different actions and observing the resulting rewards). The goal of reinforcement learning algorithms in general is to eventually stabilize (converge) to a strategy that maximizes the agent's payoff. Traditional reinforcement learning algorithms (such as Q-learning) guarantee convergence to the optimal policy in a *stationary environment* (Sutton & Barto, 1999), which simply means that the reward distribution associated with each action is fixed and does not change over time.

In a multiagent system, the reward each agent receives for executing a particular action depends on other agents' actions as well. For example, consider extending the MAB problem to the multiagent case. The reward agent A gets for choosing lever 1 depends on which lever agent B has chosen. If both agents A and B are learning and adapting their strategies, the stationary assumption of the single-agent case is violated (the reward distribution is changing) and therefore single-agent reinforcement learning techniques are not guaranteed to converge. Furthermore, in a multi-agent context the optimality criterion is not as clear as in the single agent case. Ideally, we want all agents to reach the equilibrium that maximizes their individual payoffs. However, when agents do not communicate and/or agents are not cooperative, reaching a globally optimal equilibrium is not always attainable (Claus & Boutilier, 1998). An alternative goal that we pursue here is converging to the Nash Equilibrium (NE) (Bowling, 2005; Conitzer & Sandholm, 2007; Banerjee & Peng, 2007), which is by definition a local maximum across agents (no agent can do better by deviating unilaterally from the NE).

An important aspect of understanding the behavior of a MARL algorithm is analyzing the dynamics of the algorithm: how policies of multiple learning agents evolve over time while interacting with one another. Such an analysis not only reveals whether agents using a particular MARL algorithm will eventually converge, but also points out features of the MARL algorithm that are exhibited during the convergence period. Analyzing the dynamics of a MARL algorithm in even the simplest domains (particularly, two-player-two-action games) is a challenging task and therefore was performed on only few, relatively simple, MARL algorithms (Singh, Kearns, & Mansour, 2000; Bowling & Veloso, 2002) and for a restrictive subset of games. These analyzed dynamics were either linear or piece-wise linear.

Recently, several multi-agent reinforcement learning (MARL) algorithms have been proposed and studied (Claus & Boutilier, 1998; Singh et al., 2000; Peshkin, Kim, Meuleau, & Kaelbling, 2000; Littman, 2001; Bowling & Veloso, 2002; Hu & Wellman, 2003; Bowling, 2005; Abdallah & Lesser, 2006; Conitzer & Sandholm, 2007; Banerjee & Peng, 2007). Most of the MARL algorithms assumed an agent knew the underlying game structure, or the Nash Equilibrium (NE) (Bowling & Veloso, 2002; Banerjee & Peng, 2007). Some even required knowing what actions other agents executed and what rewards they received (Hu & Wellman, 2003; Conitzer & Sandholm, 2007). These assumptions are restrictive in open domains with limited communication (such as ebay or Peer-to-Peer file-sharing) where an agent rarely knows of other agents' existence let alone observing their actions and knowing how their actions affect the agent's local reward.

On the other hand, if agents are unaware of the underlying game (particularly the NE) and they are not observing each other, then even a simple game with two players and two actions can be challenging. For example, suppose Player 1 observes that its getting reward 10 from executing its first action and reward 8 for executing action 2. As time passes, and Player 2 changes its policy, Player 1 observes a switch in the reward associated with each action: action 1 now has a reward of 7 and action 2 has a reward of 9. Note that in both cases Player 1 is unaware of its own NE strategy and is oblivious to the current policy of Player 2. The only feedback Player 1 is getting is the change in its reward function, which in turn depends on Player 2's policy. The same situation applies in reverse to Player 2.

In this paper we propose a new MARL algorithm that enables agents to converge to a Nash Equilibrium, in benchmark games, assuming each agent is oblivious to other agents and receives only one type of feedback: the reward associated with choosing a given action. The new algorithm is called the Weighted Policy Learner or WPL for reasons that will become clear shortly. We experimentally show that WPL converges in well-known benchmark two-player-two-action games. Furthermore, we show that WPL converges in the challenging Shapley's game, where state-of-the-art MARL algorithms failed to converge (Bowling & Veloso, 2002; Bowling, 2005),[1] unlike WPL. We also show that WPL outperforms state-of-the-art MARL algorithms (shorter time to converge, better performance during convergence, and better average reward) in a more realistic domain of 100 agents interacting and learning with one another. We also analyze WPL's dynamics and show that they are non-linear. This non-linearity made our attempt to solve, symbolically, the differential equations representing WPL's dynamics unsuccessful. Instead, we solve the equations numerically and verify that our theoretical analysis and our experimental results are consistent. We compare the dynamics of WPL with earlier MARL algorithms and discuss interesting differences and similarities.

The paper is organized as follows. Section 2 lays out the necessary background, including game theory and closely related MARL algorithms. Section 3 describes our proposed algorithm. Section 4 analyzes WPL's dynamics for this restricted class of games and compares it to previous algorithms' dynamics. Section 5 discusses our experimental results. We conclude in Section 6.

## 2. Background

In this section we introduce the necessary background for our contribution. First, a brief review of relevant game theory definitions is given. Then a review of relevant MARL algorithms is provided, with particular focus on gradient-ascent MARL (GA-MARL) algorithms that are closely related to our algorithm.

### 2.1 Game Theory

Game theory provides a framework for modeling agents' interaction and was used by previous researchers in order to analyze the convergence properties of MARL algorithms (Claus & Boutilier, 1998; Singh et al., 2000; Bowling & Veloso, 2002; Wang & Sandholm, 2003; Bowling, 2005; Conitzer & Sandholm, 2007; Abdallah & Lesser, 2006). A game specifies, in a compact and simple manner, how the payoff of an agent depends on other agents' actions. A (normal form) game is defined by the tuple $\langle n, A_1, ..., A_n, R_1, ..., R_n \rangle$, where $n$ is the number of players[2] in the game, $A_i$ is the set of actions available to agent $i$, and $R_i : A_1 \times ... \times A_n \to \Re$ is the reward (payoff) agent $i$ will receive as a function of the joint action executed by all agents. If the game has only two players, then it is convenient to define their reward functions as a payoff matrix as shown in Table 1. Each cell $(i, j)$ in the matrix represents the payoff received by the row player (Player 1) and the column player (Player 2), respectively, if the row player plays action $i$ and the column player plays action $j$. Table 1 and Table 2 provide example benchmark games that were used in evaluating previous MARL algorithms (Bowling & Veloso, 2002; Bowling, 2005; Conitzer & Sandholm, 2007; Banerjee & Peng, 2007).

---

1. Except for MARL algorithms that assumed agents know information about the underlying game (Hu & Wellman, 2003; Conitzer & Sandholm, 2007; Banerjee & Peng, 2007).

2. We use the terms agent and player interchangeably.

Table 1: Benchmark 2-action games. The *coordination* game has two pure NE(s): $[(0,1)_r, (0,1)_c]$ and $[(1,0)_r, (1,0)_c]$. Both *matching-pennies* and *tricky* games have one mixed NE where all actions are played with equal probability, NE=$[(\frac{1}{2}, \frac{1}{2})_r, (\frac{1}{2}, \frac{1}{2})_c]$

(a) coordination

|    | a1  | a2  |
|----|-----|-----|
| a1 | 2,1 | 0,0 |
| a2 | 0,0 | 1,2 |

(b) matching pennies

|   | H    | T    |
|---|------|------|
| H | 1,-1 | -1,1 |
| T | -1,1 | 1,-1 |

(c) tricky

|    | a1  | a2  |
|----|-----|-----|
| a1 | 0,3 | 3,2 |
| a2 | 1,0 | 2,1 |

(d) general 2-player-2-action game

|    | a1            | a2            |
|----|---------------|---------------|
| a1 | $r_{11}, c_{11}$ | $r_{12}, c_{12}$ |
| a2 | $r_{21}, c_{21}$ | $r_{22}, c_{22}$ |

Table 2: Benchmark 3-action games. Both games have one mixed NE where all actions are played with equal probability, NE= $[(\frac{1}{3}, \frac{1}{3}, \frac{1}{3})_r, (\frac{1}{3}, \frac{1}{3}, \frac{1}{3})_c]$.

(a) rock paper scissors

|    | c1   | c2   | c3   |
|----|------|------|------|
| r1 | 0,0  | -1,1 | 1, -1 |
| r2 | 1,-1 | 0,0  | -1,1 |
| r3 | -1,1 | 1,-1 | 0,0  |

(b) Shapley's

|    | c1  | c2  | c3  |
|----|-----|-----|-----|
| r1 | 0,0 | 1,0 | 0,1 |
| r2 | 0,1 | 0,0 | 1,0 |
| r3 | 1,0 | 0,1 | 0,0 |

A *policy* (or a *strategy*) of an agent $i$ is denoted by $\pi_i \in PD(A_i)$, where $PD(A_i)$ is the set of probability distributions over actions $A_i$. The probability of choosing an action $a_k$ according to policy $\pi_i$ is $\pi_i(a_k)$. A policy is *deterministic* or *pure* if the probability of playing one action is 1 while the probability of playing other actions is 0, (i.e. $\exists k : \pi_i(a_k) = 1$ AND $\forall l \neq k : \pi_i(a_l) = 0$), otherwise the policy is *stochastic* or *mixed*.

A *joint policy* $\pi$ is the collection of individual agents' policies, i.e. $\pi = \langle \pi_1, \pi_2, ..., \pi_n \rangle$, where $n$ is the number of agents. For convenience, the joint policy is usually expressed as $\pi = \langle \pi_i, \pi_{-i} \rangle$, where $\pi_{-i}$ is the collection of all policies of agents other than agent $i$.

Let variable $A_{-i} = \{\langle a_1, ..., a_n \rangle : a_j \in A_j \bigwedge i \neq j\}$. The *expected reward* agent $i$ would get, if agents follow a joint policy $\pi$, is $V_i(\langle \pi_i, \pi_{-i} \rangle) = \sum_{a_i \in A_i} \sum_{a_{-i} \in A_{-i}} \pi_i(a_i) \pi_{-i}(a_{-i}).R_i(a_i, a_{-i})$, i.e. the reward averaged over the joint policy. A joint policy is a *Nash Equilibrium*, or NE, iff no agent can get higher expected reward by changing its policy unilaterally. More formally, $\langle \pi_i^*, \pi_{-i}^* \rangle$ is an NE iff $\forall i : V_i(\langle \pi_i^*, \pi_{-i}^* \rangle) \geq V_i(\langle \pi_i, \pi_{-i}^* \rangle)$. An NE is *pure* if all its constituting policies are pure. Otherwise the NE is called mixed or stochastic. Any game has at least one Nash equilibrium, but may not have any pure (deterministic) equilibrium.

Consider again the benchmark games in Table 1. The *coordination* game (Table 1(a)) is an example of games that have at least one pure NE. The *matching pennies* game (Table 1(b)) is an example of games that do not have any pure NE and only have a mixed NE (where each player plays a1 and a2 with equal probability). Convergence of GA-MARL algorithms in games with pure

NE is easier than games where the only NE is mixed (Singh et al., 2000; Bowling & Veloso, 2002; Zinkevich, 2003). The tricky game is similar to matching pennies game in that it only has one mixed NE (no pure NE), yet some GA-MARL algorithms succeeded in converging in the matching pennies games, while failing in the tricky game (Bowling & Veloso, 2002).

Table 2 shows well-known 2-player-3-action benchmark games. Shapley's game, in particular, has received considerable attention from MARL community (Bowling, 2005; Conitzer & Sandholm, 2007; Banerjee & Peng, 2007) as it remains challenging for state-of-the-art MARL algorithms despite its apparent simplicity (and similarity to the rock-paper-scissors game which is not as challenging). Our proposed algorithm is the first MARL algorithm to converge in Shapley's game without observing other agents or knowing the underlying NE strategy.

Before introducing MARL algorithms in the following section, two issues are worth noting. The first issue is the general assumption in the MARL context: agents play the same game repeatedly for a large number of times. This is a necessary assumption in order for agents to gain experience and learn. The second issue is that the games' rewards, described in this section, are deterministic given the joint action. However, from each agent's perspective, the rewards are stochastic because of the randomness caused by the other agents' actions in the system.

## 2.2 Multiagent Reinforcement Learning, MARL

Early MARL algorithms were based on the Q-learning algorithm (Sutton & Barto, 1999), and therefore could only learn a deterministic policy (Claus & Boutilier, 1998), significantly limiting their applicability in competitive and partially observable domains. The Joint Action Learner (Claus & Boutilier, 1998) is an example of this family of algorithms that also required the knowledge of other agents' chosen actions.

Another class of MARL algorithms is the class of Equilibrium learners such as Nash-Q (Hu & Wellman, 2003), OAL (Wang & Sandholm, 2003), and AWESOME (Conitzer & Sandholm, 2007). Most of these algorithms assumed that each agent observed other agents' actions in addition to knowing the underlying game.[3] Each agent then computed the Nash Equilibria. The purpose of learning is to allow agents to converge to a particular Nash Equilibrium (in case all agents execute the same MARL algorithm).

Observing other agents or knowing the underlying game structure are not applicable in open and large domains, which motivated the development of gradient ascent learners. Gradient ascent MARL algorithms (GA-MARL) learn a stochastic policy by directly following the expected reward gradient. The ability to learn a stochastic policy is particularly important when the world is not fully observable or has a competitive nature. Consider for example a blind robot in a maze (i.e. the robot cannot distinguish between maze locations). Any deterministic policy (i.e. one that always chooses a particular action everywhere) may never escape the maze, while a stochastic policy that chooses each action with non-zero probability will eventually escape the maze.[4] Similarly, in a competitive domain a stochastic policy may be the only stable policy (e.g. the Nash Equilibrium in a competitive game). The remainder of this section focuses on this family of algorithms as it is closely related to our proposed algorithm.

---

3. Nash-Q did not require knowing the underlying game but required observing other agents rewards in addition to their chosen actions.

4. Note that a uniformly random policy may not be optimal if the maze is biased in a specific direction.

The Infinitesimal Gradient Ascent algorithm (IGA) (Singh et al., 2000) and its generalization (Generalized IGA or GIGA) (Zinkevich, 2003) were proved to converge in games with pure NE. However, both algorithms failed to converge in games with mixed NE, and therefore may not be suitable for applications that require a mixed policy.

Several modifications to IGA and GIGA were proposed to avoid divergence in games with mixed NE, including: IGA/PHC-WoLF (Bowling & Veloso, 2002), PHC-PDWoLF (Banerjee & Peng, 2003), and GIGA-WoLF (Bowling, 2005). They all used some form of the Win or Learn Fast heuristics (Bowling & Veloso, 2002), whose purpose is to speedup learning if the agent is doing worse than its NE policy (losing) and to slow down learning if the agent is doing better than the NE policy. The main problem with this heuristic is that an agent cannot know whether it is doing better or worse than its NE policy without knowing the underlying game a prior. Therefore, a practical implementation of the WoLF heuristic needed to use approximation methods for predicting the performance of the agent's NE policy. Our algorithm, called WPL, uses a different heuristic that does not require knowing the NE policy and consequently, as we show, converges in both the tricky game and Shapley's game, where algorithms relying on the WoLF heuristic failed. Furthermore, we show that in a large-scale partially-observable domain (the distributed task allocation problem, DTAP) WPL outperforms state-of-the-art GA-MARL algorithm GIGA-WoLF.

The following section reviews in further detail the well known GA-MARL algorithms IGA, IGA-WoLF, and GIGA-WoLF which we will use to compare our algorithm against.

## 2.3 Gradient-Ascent MARL Algorithms

The first GA-MARL algorithm whose dynamics were analyzed is the Infinitesimal Gradient Ascent (IGA) (Singh et al., 2000). IGA is a simple gradient ascent algorithm where each agent $i$ updates its policy $\pi_i$ to follow the gradient of expected payoffs (or the value function) $V_i$, as illustrated by the following equations.

$$\Delta \pi_i^{t+1} \leftarrow \eta \frac{\partial V_i(\pi^t)}{\partial \pi_i}$$

$$\pi_i^{t+1} \leftarrow projection(\pi_i^t + \Delta \pi_i^{t+1})$$

Parameter $\eta$ is called the policy-learning-rate and approaches zero in the limit ($\eta \rightarrow 0$), hence the word Infinitesimal in IGA. Function $projection$ projects the updated policy to the space of valid policies. The original IGA paper (Singh et al., 2000) defined the $projection$ function (and therefore IGA) in case each agent has only two actions to choose from. A general definition of the $projection$ function was later developed and the resulting algorithm was called Generalized IGA or GIGA (Zinkevich, 2003). The generalized function is $projection(x) = argmin_{x':valid(x')}|x - x'|$, where $|x - x'|$ is the Euclidean distance between $x$ and $x'$. A valid policy $\pi$ (over a set of actions $A$) must satisfy two constraints: $\forall a \in A : 1 \geq \pi(a) \geq 0$ and $\sum_\pi = \sum_{a \in A} \pi(a) = 1$.

In other words, the space of valid policies is a simplex, which is a line segment (0,1),(1,0) in case of two actions, a triangular surface (1,0,0), (0,1,0), (0,0,1) in case of three actions, and so on. A joint policy, $\pi$, is a point in this simplex. It is quite possible (especially when trying to follow an approximate policy gradient) that $\sum_\pi$ deviates from 1 and goes beyond the simplex. The generalized $projection$ function projects an invalid policy to the *closest* valid policy within the simplex (Zinkevich, 2003).

The IGA algorithm did not converge in all two-player-two-action games (Singh et al., 2000). Algorithm IGA-WoLF (WoLF stands for Win or Learn Fast) was proposed (Bowling & Veloso, 2002) in order to improve convergence properties of IGA by using two different learning rates as follows. If a player is getting an average reward lower than the reward it would get for executing its NE strategy, then the learning rate should be large. Otherwise (the player current policy is better than its NE policy), the learning rate should be small. More formally,

$$\Delta \pi_i(a) \leftarrow \frac{\partial V_i(\pi)}{\partial \pi_i}(a) \cdot \left\{ \begin{array}{ll} \eta_{lose} & \text{if } V_i(\pi_i, \pi_{-i}) < V_i(\pi_i^*, \pi_{-i}) \\ \eta_{win} & \text{otherwise} \end{array} \right.$$

$$\pi_i \leftarrow projection(\pi_i + \Delta \pi_i)$$

Where $\pi_i^*$ is the NE policy for agent $i$ and $\eta_{lose} > \eta_{win}$ are the learning rates. The dynamics of IGA-WoLF have been analyzed and proved to converge in all 2-player-2-action games (Bowling & Veloso, 2002), as we briefly review Section 2.4. IGA-WoLF has limited practical use, because it required each agent to know its equilibrium policy (which means knowing the underlying game). An approximation to IGA-WoLF was proposed, PHC-WoLF, where an agent's NE strategy is approximated by averaging the agent's own policy over time (Bowling & Veloso, 2002). The approximate algorithm, however, failed to converge in the *tricky* game shown in Table 1(c).

The GIGA-WoLF algorithm extended the GIGA algorithm with the WoLF heuristic. GIGA-WoLF kept track of two policies $\pi$ and $z$. Policy $\pi$ was used to select actions for execution. Policy $z$ was used to approximate the NE policy. The update equations of GIGA-WoLF are (Bowling, 2005, 2004):

$$\hat{\pi}^{t+1} = projection(\pi^t + \delta r^t) \tag{1}$$

$$z^{t+1} = projection(\pi^t + \delta r^t/3) \tag{2}$$

$$\eta^{t+1} = \min\left(1, \frac{||z^{t+1} - z^t||}{z^{t+1} - \hat{\pi}^t}\right) \tag{3}$$

$$\pi^{t+1} = \hat{\pi}^{t+1} + \eta^{t+1}(z^{t+1} - \hat{\pi}^{t+1}) \tag{4}$$

The main idea is a variant of the WoLF heuristics (Bowling & Veloso, 2002). An agent $i$ changes its policy $\pi_i$ faster if the agent is performing worse than policy $z$, i.e. $V^{\pi_i} < V^z$. Because $z$ moves slower than $\pi$, GIGA-WoLF uses $z$ to realize that it needs to change the current gradient direction. This approximation allows GIGA-WoLF to converge in the tricky game, but not in Shapley's game (Bowling, 2005).

The following section briefly reviews the analysis of IGA's and IGA-WoLF's dynamics, showing how a joint policy of two agents evolves over time. We will then build on this analysis in Section 4 when we analyze WPL's dynamics.

## 2.4 Dynamics of GA-MARL Algorithms

Differential equations were used to model the dynamics of IGA (Singh et al., 2000). To simplify analysis, the authors only considered two-player-two-action games, as we also do here. We will refer to the joint policy of the two players at time $t$ by the probabilities of choosing the first action $(p^t, q^t)$, where $\pi_1 = (p^t, 1 - p^t)$ is the policy of player 1 and $\pi_2 = (q^t, 1 - q^t)$ is the policy of player 2. The $t$ notation will be omitted when it does not affect clarity (for example, when we are considering only one point in time).

IGA's update equations can be simplified to be (note that $r_{ij}$ and $c_{ij}$ are the payoffs for row and column players respectively):

$$p^{t+1} = p^t + \eta \frac{\partial V_r(p^t, q^t)}{\partial p} = p^t + \eta (V_r(1, q^t) - V_r(0, q^t))$$

where

$$\begin{aligned}
V_r(1, q^t) - V_r(0, q^t) &= \left(r_{11}q^t + r_{12}(1 - q^t)\right) - \left(r_{21}q^t + r_{22}(1 - q^t)\right) \\
&= q^t(r_{11} - r_{12} - r_{21} + r_{22}) + (r_{12} - r_{22}) \\
&= u_1 q^t + u_2
\end{aligned} \tag{5}$$

and similarly,

$$q^{t+1} = q^t + \eta \cdot (u_3 p^t + u_4)$$

where $u_1, u_2, u_3,$ and $u_4$ are game-dependent constants having the following values.

$$u_1 = r_{11} - r_{12} - r_{21} + r_{22}$$
$$u_2 = r_{12} - r_{22}$$
$$u_3 = c_{11} - c_{12} - c_{21} + c_{22}$$
$$u_4 = c_{21} - c_{22}$$

In analyzing IGA, the original paper distinguished between three types of games depending on the $u_1$ and $u_3$ parameters: $u_1 u_3 > 0$, $u_1 u_3 = 0$, and $u_1 u_3 < 0$. Figure 1(a), Figure 1(b), and Figure 1(c) are taken from (Bowling & Veloso, 2002) and illustrate the dynamics of IGA for each of the three cases. Each figure displays the space of joint policies, where the horizontal axis represent the policy of the row player, $p$, and the vertical axis represents the policy of the column player $q$.[5]

The joint policy of the two agents evolves over time by following the directed lines in the figures. For example, Figure 1(b) illustrates that starting from any joint policy, the two players will eventually evolve to one of two equilibriums, either the bottom-left corner, or the upper-right corner. It should be noted that, for simplicity, the dynamics shown in Figure 1 are unconstrained: the effect of the *projection* function (the simplex described in Section 2.3) is not shown. In IGA's analysis (Singh et al., 2000), the effect of the *projection* function was taken into account by considering all possible locations of the simplex.

In the first and the second cases, IGA converged. In the third case IGA oscillated around the equilibrium strategy, without ever converging to it. This happens because at any point of time, unless both agents are at the NE, one agent is better off changing its policy to be closer to its NE strategy, while the other agent is better off changing its policy to be further away from its NE strategy. These roles switch as one of the agents crosses its NE strategy as shown in Figure 1(c).

As described in Section 2.3, IGA-WoLF was proposed (Bowling & Veloso, 2002) as an extension to IGA that ensures convergence in the third case. The idea of IGA-WoLF, as described earlier, is to have two learning rates by which an agent is moving toward or away from the NE. The following equations capture the dynamics of IGA-WoLF, where $\langle p^*, q^* \rangle$ is the NE. The equations use a factored form of $\eta_{lose}$ and $\eta_{win}$.

---

5. Note that Figure 1(b) and Figure 1(c) are divided into four quadrants A, B, C, and D for clarification. The gradient direction in each quadrant is illustrated by arrows in Figure 1(c).

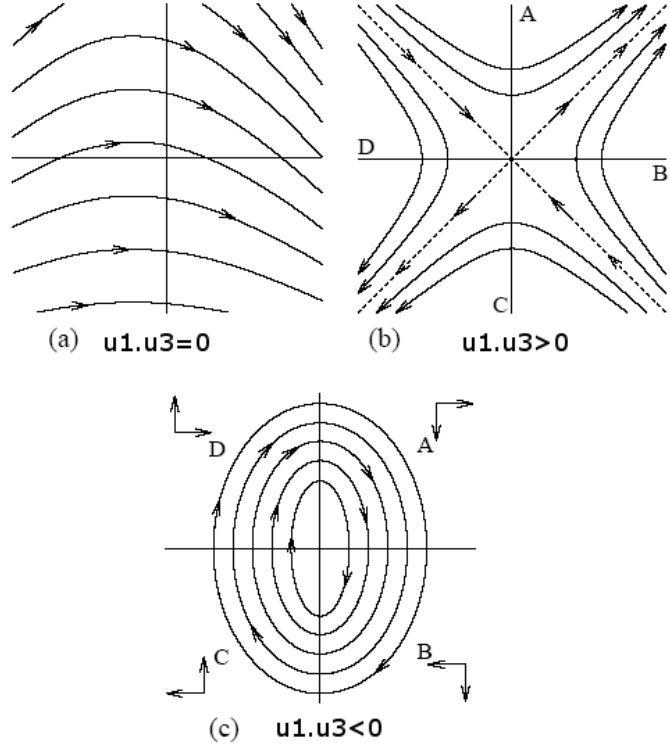

Figure 1: An illustration of IGA's dynamics (Bowling & Veloso, 2002).

$$p^t = p^{t-1} + \eta(u_1 q^{t-1} + u_2) \cdot \begin{cases} l_{lose} & \text{if } V_r(p^{t-1}, q^{t-1}) < V_r(p^*, q^{t-1}) \\ l_{win} & \text{otherwise} \end{cases}$$

$$q^t = q^{t-1} + \eta(u_3 p^{t-1} + u_4) \cdot \begin{cases} l_{lose} & \text{if } V_c(q^{t-1}, p^{t-1}) < V_c(q^*, p^{t-1}) \\ l_{win} & \text{otherwise} \end{cases}$$

The dynamics of IGA-WoLF is best illustrated visually by Figure 2, which is taken from (Bowling & Veloso, 2002). The switching between learning rates causes switching between ellipses of smaller diameters, eventually leading to convergence.

The main limitation of IGA-WoLF is that it assumed each agent knows the NE policy (needed to switch between the two modes of the learning rate). The following section presents WPL, which does not make this assumption, at the expense of having more complex dynamics as we show in Section 4.

## 3. Weighted Policy Learner (WPL)

We propose in this paper the Weighted Policy Learner (WPL) algorithm, which has the following update equations:

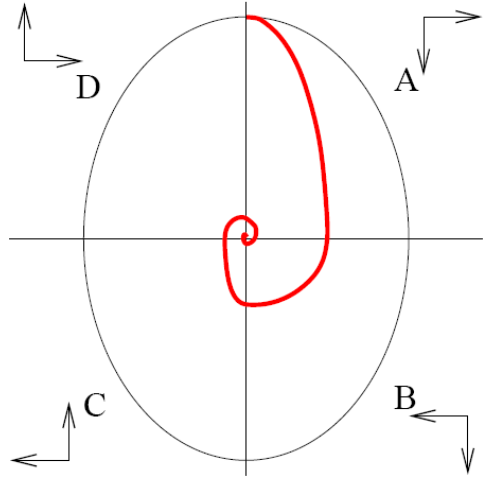

Figure 2: An illustration of IGA-WoLF's dynamics for the case of mixed NE (Bowling & Veloso, 2002).

$$\Delta\pi_i(a) \leftarrow \frac{\partial V_i(\pi)}{\partial \pi_i(a)} \cdot \eta \cdot \begin{cases} \pi_i(a) & \text{if } \frac{\partial V_i(\pi)}{\partial \pi_i(a)} < 0 \\ 1 - \pi_i(a) & \text{otherwise} \end{cases}$$

$$\pi_i \leftarrow projection(\pi_i + \Delta\pi_i)$$

The $projection$ function is adopted from GIGA (Zinkevich, 2003) with minor modification: $\forall a : 1 \geq \pi(a) \geq \epsilon$ (the modification ensures a minimum amount of exploration $\epsilon$). The algorithm works as follows. If the gradient for a particular action is negative then the gradient is weighted by $\pi_i(a)$, otherwise (gradient is positive) the gradient is weighted by $(1-\pi_i(a))$. So effectively, the probability of choosing a good action increases by a rate that decreases as the probability approaches 1 (or the boundary of the simplex). Similarly, the probability of choosing a bad action decreases at a rate that also decreases as the probability approaches zero.

So unless the gradient direction changes, a WPL agent decreases its learning rate as the agent gets closer to simplex boundary. This means, for a 2-player-2-action game, a WPL agent will move towards its NE strategy (away from the simplex boundary) faster than moving away from its NE strategy (towards the simplex boundary), because the NE strategy is always inside the simplex.

WPL is biased against pure (deterministic) policies and will reach a pure policy only in the limit (because the rate of updating the policy approaches zero). This theoretical limitation is of little practical concern for two reasons. The first reason is exploration: 100% pure strategies are bad because they prevent agents from exploring other actions (in an open dynamic environment the reward of an action may change over time).

The second reason is that if an action dominates another action (the case when a game has a pure NE), then $\frac{\partial V_i(\pi)}{\partial \pi_i(a)}$ can be large enough that $\frac{\partial V_i(\pi)}{\partial \pi_i(a)} \cdot \eta > 1$. In this case WPL will jump to a pure policy in one step.[6] Note that in Section 4, in order to simplify theoretical analysis, we assume $\eta \to 0$. From a practical perspective, however, $\eta$ is set to value close to 0, but never 0. This holds

---

6. In fact, WPL can even go beyond the simplex of valid policies, that is when the $projection$ function comes into play

for all gradient-ascent MARL algorithms (Bowling & Veloso, 2002; Bowling, 2005; Banerjee & Peng, 2007).

There are key differences between IGA-WoLF and WPL's update rules despite the apparent similarity. Both algorithms have two modes of learning rates (corresponding to the two conditions in the policy update rule). However, IGA-WoLF needs to know the equilibrium strategy in order to distinguish between the two modes, unlike WPL that needs to know only the value of each action. Another difference is that IGA-WoLF used two fixed learning rates ($\eta_l > \eta_w$) for the two modes, while WPL uses a continuous spectrum of learning rates, depending on the current policy. This can be understood from the definition of $\Delta\pi_i$ in WPL's update equations, which includes an additional (continuous) scaling factor: $\pi_i^t$. This particular feature causes WPL's dynamics to be non-linear, as we discuss in the following section.

## 4. Analyzing WPL's Dynamics

**Claim 1** *WPL has non-linear dynamics.*

**Proof.**

The policies of two agents following WPL can be expressed as follows

$$q^t \leftarrow q^{t-1} + \eta(u_3 p^{t-1} + u_4) \begin{cases} 1 - q^{t-1} & \text{if } u_3 p^{t-1} + u_4 > 0 \\ q^{t-1} & \text{otherwise} \end{cases}$$

and

$$p^t \leftarrow p^{t-1} + \eta(u_1 q^{t-1} + u_2) \begin{cases} 1 - p^{t-1} & \text{if } u_1 q^{t-1} + u_2 > 0 \\ p^{t-1} & \text{otherwise} \end{cases}$$

Note that $u_1 q^{t-1} + u_2 = V_r(1, q^{t-1}) - V_r(0, q^{t-1})$ from Equation 5. It then follows that

$$\frac{p^t - p^{t-1}}{\eta\left(t - (t-1)\right)} =$$

$$(u_1 q^{t-1} + u_2) \begin{cases} 1 - p^{t-1} & \text{if } V_r(1, q^{t-1}) > V_r(0, q^{t-1}) \\ p^{t-1} & \text{otherwise} \end{cases}$$

and analogously for the column player

$$\frac{q^t - q^{t-1}}{\eta\left(t - (t-1)\right)} =$$

$$(u_3 p^{t-1} + u_4) \begin{cases} 1 - q^{t-1} & \text{if } V_c(1, p^{t-1}) > V_c(0, p^{t-1}) \\ q^{t-1} & \text{otherwise} \end{cases}$$

As $\eta \to 0$, the equations above become differential:

$$q'(t) =$$

$$(u_3 p^{t-1} + u_4) \begin{cases} 1 - q^{t-1} & \text{if } V_c(1, p^{t-1}) > V_c(0, p^{t-1}) \\ q^{t-1} & \text{otherwise} \end{cases} \tag{6}$$

$$p'(t) =$$
$$(u_1 q^{t-1} + u_2) \begin{cases} 1 - p^{t-1} & \text{if } V_r(1, q^{t-1}) > V_r(0, q^{t-1}) \\ p^{t-1} & \text{otherwise} \end{cases} \tag{7}$$

Notice here that the NE strategy does not appear in WPL's equations, unlike IGA-WoLF. Furthermore, while IGA and IGA-WoLF needed to take the *projection* function into account, we can safely ignore the *projection* function while analyzing the dynamics of WPL for two-player-two-action games. This is due to the way WPL scales the learning rate using the current policy. By the definition of $p'(t)$, a positive $p'(t)$ approaches zero as $p$ approaches one and a negative $p'(t)$ approaches zero as $p$ approaches zero. In other words, as $p$ (or $q$) approaches 0 or 1, the learning rate approaches zero, and therefore $p$ (or $q$) will never go beyond the simplex of valid policies.[7] This observation will become apparent in Section 4.2 when we compare the dynamics of WPL to the dynamics of both IGA and IGA-WoLF.

Following IGA-WoLF's analysis (Bowling & Veloso, 2002), as illustrated in Figure 3, we will focus on the challenging case when there is no deterministic NE (the NE is inside the joint policy space) and analyze how $p$ and $q$ evolve over time. This is the case where the original IGA oscillated as shown in Figure 1. It is important to note, however, that all gradient ascent MARL algorithms (including WPL) converge in 2x2 games cases where there is (at least) one pure NE, because the dynamics do not have any loops and eventually lead to one of the pure equilibriums (Singh et al., 2000; Bowling & Veloso, 2002).

We will solve WPL's differential equations for the period $0 \rightarrow T4$ in Figure 3, assuming that Player 2 starts at its NE policy $q^*$ at time 0 and returns to $q^*$ at time $T4$. If we can prove that, over the period $0 \rightarrow T4$, Player 2's policy $p(t)$ gets closer to the NE policy $p^*$, i.e. $p_{min2} - p_{min1} > 0$ in Figure 3, then by induction the next time period $p$ will get closer to the equilibrium and so on.

For readability, $p$ and $q$ will be used instead of $p(t)$ and $q(t)$ for the remainder of this section. Without loss of generality, we can assume that $u_1 q + u_2 > 0$ iff $q > q^*$ and $u_3 p + u_4 > 0$ iff $p < p^*$. The overall period $0 \rightarrow T4$ is divided into four intervals defined by times $0, T1, T2, T3$, and $T4$. Each period corresponds to one combination of $p$ and $q$ as follows. For the first period $0 \rightarrow T1$, $p < p^*$ and $q > q^*$, therefore agent $p$ is better off moving toward the NE, while agent $q$ is better off moving away from the NE. The differential equations can be solved by dividing $p'$ and $q'$

$$\frac{dp}{dq} = \frac{(1-p)(u_1 q + u_2)}{(1-q)(u_3 p + u_4)}$$

Then by separation we have

$$\int_{p_{min1}}^{p^*} \frac{u_3 p + u_4}{1 - p} dp = \int_{q^*}^{q_{max}} \frac{u_1 q + u_2}{1 - q} dq$$

$$-u_3(p^* - p_{min1}) + (u_3 + u_4) ln \frac{1 - p_{min1}}{1 - p^*} =$$
$$-u_1(q_{max} - q^*) + (u_1 + u_2) ln \frac{1 - q^*}{1 - q_{max}}$$

---

7. In practice WPL still needs the *projection* function because $\eta$ must be set to a small value strictly larger than 0.

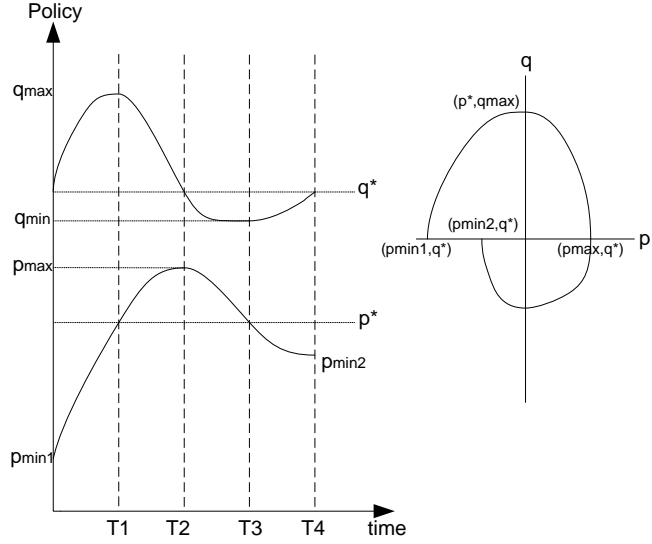

Figure 3: An illustration of WPL's dynamics. The figure to the left shows policies evolution over time, while the figure to the right shows the joint policy space.

Similarly, for period $T1 \rightarrow T2$, where $p > p^*$ and $q > q^*$

$$-u_3(p_{max} - p^*) + (u_3 + u_4)ln\frac{1 - p^*}{1 - p_{max}} =$$

$$u_1(q^* - q_{max}) + u_2 ln\frac{q^*}{q_{max}}$$

and for period $T2 \rightarrow T3$, where $p > p^*$ and $q < q^*$

$$u_3(p^* - p_{max}) + u_4 ln\frac{p^*}{p_{max}} = u_1(q_{min} - q^*) + u_2 ln\frac{q_{min}}{q^*}$$

and finally for period $T3 \rightarrow T4$, where $p < p^*$ and $q < q^*$

$$u_3(p_{min2} - p^*) + u_4 ln\frac{p_{min2}}{p^*} =$$

$$-u_1(q^* - q_{min}) + (u_1 + u_2)ln\frac{1 - q_{min}}{1 - q^*}$$

These are 4 *non-linear* equations (note the existence of both $x$ and $ln(x)$ in all equations) in 5 unknowns ($p_{min}, p_{min2}, p_{max}, q_{min1}, q_{max}$), along with the inequalities governing the constants $u_1, u_2, u_3$, and $u_4$.

Because WPL's dynamics are non-linear, we could not obtain a closed-form solution and therefore we could not formally prove WPL's convergence.[8] Instead, we solve the equations numerically in the following section. Although a numerical solution is still not a formal proof, it provides useful insights into understanding WPL's dynamics.

---

8. If the equations were linear, we could have substituted for all unknowns in terms of $p_{min1}$ and $p2_{min}$ and easily determined whether $p_{min2} - p_{min1} > 0$, similar to IGA-WoLF.

### 4.1 Numerical Solution

We have used Matlab to solve the equations numerically. Figure 4 shows the theoretical behavior predicted by our model for the matching-pennies game. There is a clear resemblance to the actual (experimental) behavior for the same game (Figure 5). Note that the time-scale on the horizontal axes of both figures are effectively the same, because what is displayed on the horizontal axis in Figure 5 is decision steps in the simulation. When multiplied by the actual policy-learning-rate $\eta$ (the time step) used in the experiments, $0.001$, both axes become identical.

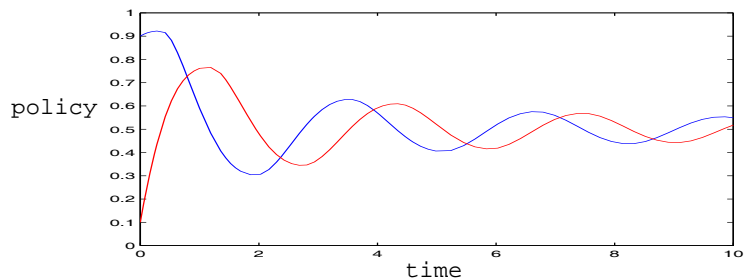

Figure 4: Convergence of WPL as predicted by the theoretical model for the matching pennies game.

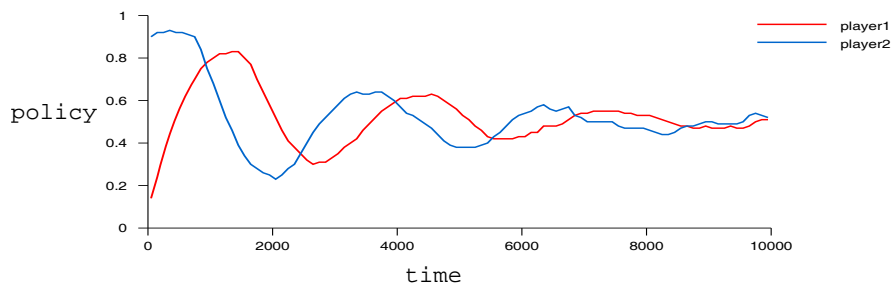

Figure 5: Convergence of WPL through experiments, using $\eta = 0.001$ .

Figure 6 plots $p(t)$ versus $q(t)$, for a game with NE= $(0.9, 0.9)$ ($u1 = 0.5, u2 = -0.45, u3 = -0.5, , u4 = 0.45$) and starting from 157 different initial joint policies (40 initial policies over each side of the joint policy space). Figure 7 plots $p(t)$ and $q(t)$ against time, showing convergence from each of the 157 initial joint policies.

We repeated the above numerical solution for 100 different NE(s) that make a 10x10 grid in the p-q space (and starting from the 157 boundary initial joint policies for each individual NE). The WPL algorithm converges to the NE in a spiral fashion similar to the specific case in Figure 6 in all the 100 NE(s). Instead of drawing 100 figures (one for each NE), Figure 8 plots the merge of the 100 figures in a compact way: plotting agents' joint policy from time 700 to time 800 (which is enough for convergence as Figure 7 shows). The two agents converge in all the 100 NE cases, as indicated by the centric points. Note that if the algorithm was not converging, then the joint policies trajectories after time 700 should occupy more space, because the 157 initial joint policies are on the policy space boundary.

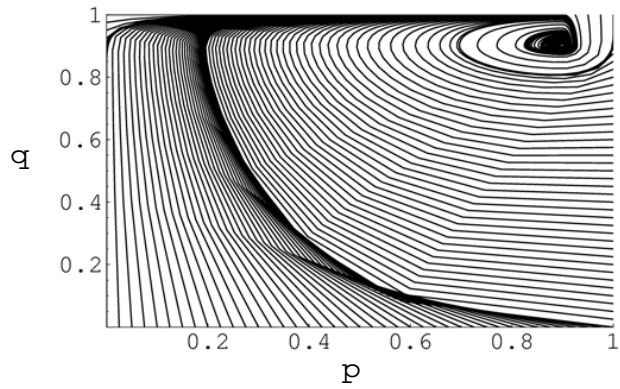

Figure 6: An illustration of WPL convergence to the (0.9,0.9) NE in the p-q space: p on the horizontal axis and q on the vertical axis.

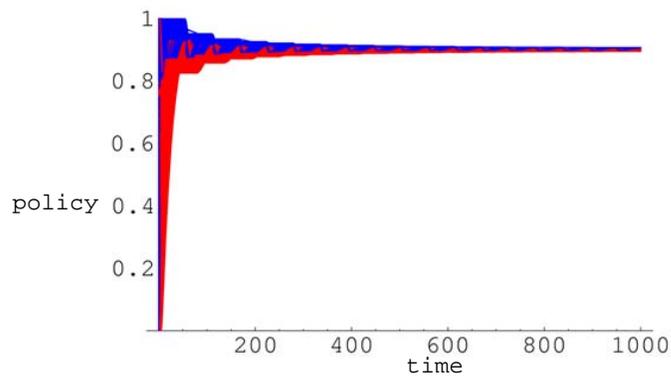

Figure 7: An illustration of WPL convergence to the (0.9,0.9) NE: p(t) (gray) and q(t) (black) are plotted on the vertical axis against time (horizontal axis).

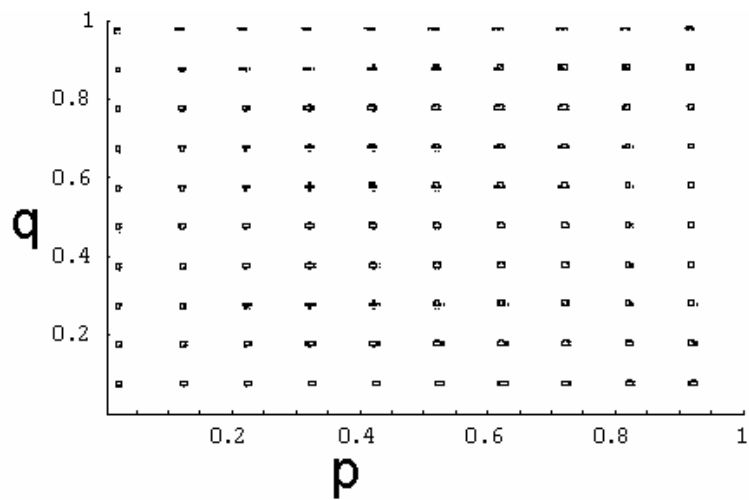

Figure 8: An illustration of WPL's convergence for 10x10 NE(s).

### 4.2 Comparing Dynamics of IGA, IGA-WoLF, and WPL

With differential equations modeling each of the three algorithms, we now compare their dynamics and point out the main distinguishing characteristics of WPL. Matlab was again used to solve the differential equations (of the three algorithms) numerically. Figure 9 shows the dynamics for a game with $u1u3 < 0$ and the NE=(0.5,0.5). The joint strategy moves in clockwise direction. The dynamics of WPL are very close to IGA-WoLF, with IGA-WoLF converging a bit faster (after one complete round around the NE, IGA-WoLF is closer to the NE than WPL). It is still impressive that WPL has comparable performance to IGA-WoLF, since WPL does not require agents to know their NE strategy or the underlying game a priori, unlike IGA-WoLF.

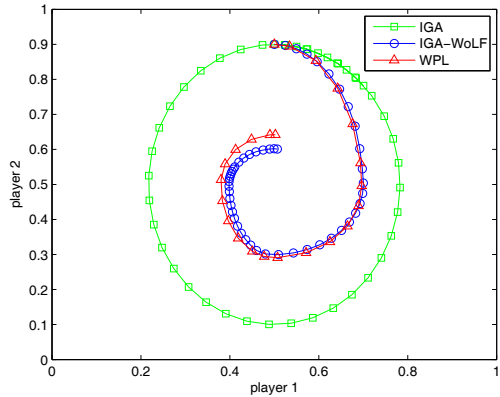

Figure 9: Dynamics of IGA, IGA-WoLF, and WPL in a game with NE=(0.5,0.5).

Figure 10 shows the dynamics in a game with $u1u3 < 0$ but the NE=(0.5,0.1). Three interesting regions in the figure are designated with A,B, and C. Region A shows that both IGA and IGA-WoLF dynamics are *discontinuous* due to the effect of the *projection* function. Because WPL uses a smooth policy weighting scheme, the dynamics remain continuous. This is also true in region B. In region C, WPL initially deviates from the NE more than IGA, but eventually converges as well. The reason is that because the NE, in this case, is closer to the boundary, policy weighting makes the vertical player move at a much slower pace when moving downward (the right half) than the horizontal player.

Figure 11 shows the dynamics for the coordination game (Table 1(a)), starting from initial joint policy (0.1,0.6). The coordination game has two NEs: (0,0) and (1,1). All algorithms converge to the closer NE, (0,0), but again we see that both IGA and IGA-WoLF have discontinuity in their dynamics, unlike WPL which smoothly converge to the NE. Notice that WPL converges to a pure NE in the limit, but the graph shows the joint policy space, so there is no depiction of time.

The continuity of WPL's dynamics comes into play when the target policy is mixed. While IGA, GIGA, and GIGA-WoLF algorithms go through extreme deterministic policies during the transient period prior to convergence, which can cause ripple effect in realistic settings where large number of agents are interacting asynchronously (e.g. through a network). WPL never reaches such extremes and the experimental results in Section 5.4 show that GIGA-WoLF takes significantly more time to converge compared to WPL in a domain of 100 agents.

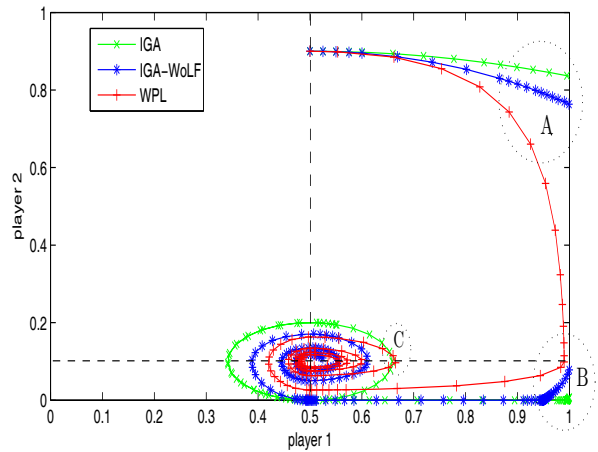

Figure 10: Dynamics of IGA, IGA-WoLF, and WPL in a game with NE=(0.5,0.1). Three regions are of particular interest and highlighted in the figure: A, B, and C.

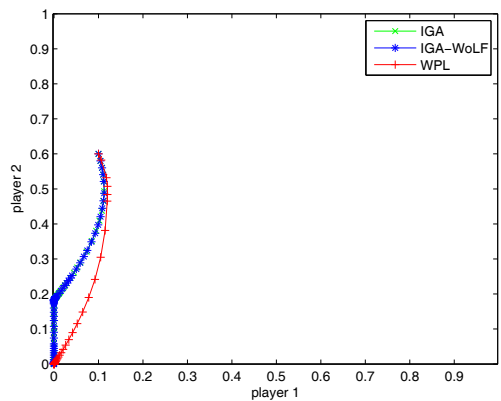

Figure 11: Dynamics of IGA, IGA-WoLF, and WPL in the coordination game with two NEs=(0,0) and (1,1). Note that IGA's dynamics is exactly as IGA-WoLF's dynamics in this case.

The following section presents our experimental results for both benchmark 2-player-2-action games and larger games involving more actions and more agents.

## 5. Experimental Results

This section is divided into three parts. Section 5.1 discusses main learning parameters that need to be set in practice. Section 5.2 presents the experimental results for the 2x2 benchmark games. Section 5.3 presents the experimental results for domains with larger number of actions and/or agents.

### 5.1 Learning Parameters

Conducting experiments for WPL involves setting two main parameters: the policy-learning-rate, $\eta$, and the value-learning-rate, $\alpha$. In theory (as discussed in Section 4) the policy-learning-rate, $\eta$, should approach zero. In practice, however, this is not possible and we have tried setting $\eta$ to different small values between 0.01 and 0.00001. The smaller $\eta$ is, the longer it will take WPL to converge, and the smaller the oscillations around the NE will become (and vice versa), similar to previous GA-MARL algorithms. A reasonable value that we have used in most of our experiments is $\eta = 0.002$.

The value-learning-rate $\alpha$ is used to compute the expected reward of an action $a$ at time $t$, or $r^t(a)$, which is not known *priori*. The common approach, which was used in previous GA-MARL algorithms and we also use here, is using the equation $r^{t+1}(a) \leftarrow \alpha R^t + (1-\alpha)r^t(a)$, where $R^t$ is the sample reward received at time $t$ and $0 \leq \alpha \leq 1$ (Sutton & Barto, 1999). We have tried three values for $\alpha$: 0.01,0.1, and 1.

### 5.2 Benchmark 2x2 Games

We have implemented a set of previous MARL algorithms for comparison: PHC-WoLF (the realistic implementation of IGA-WoLF), GIGA, and GIGA-WoLF. In our experiments we have not used any decaying rates. The reason is that in an open system where dynamics can continuously change, one would want learning to be continuous as well. We have fixed the exploration rate $\epsilon$ to 0.1 (which comes into play in the modified *projection* function in Section 3), the policy-learning-rate $\eta$ to 0.002, and the value-learning-rate $\alpha$ to 0.1 (unless otherwise specified).

The first set of experiments show the results of applying our algorithm on the three benchmark games described in Table 1 over 10 simulation runs. Figure 12 plots $\pi(r1)$ and $\pi(c1)$ (i.e. the probability of choosing the first action for the row player and the column player respectively) against time. For the matching pennies and the tricky games, the initial joint policy is $([0.1, 0.9]_r, [0.9, 0.1]_c)$ (we also tested 0.5,0.5 as initial joint policy with similar results) and the plot is the average across the 10 runs, with standard deviation shown as vertical bar. In the coordination game we plotted a single run because there are two probable pure NEs and averaging over runs will not capture that. WPL converges to one of the NE strategies in all the runs.

Figure 13 shows the convergence of GIGA, PHC-WoLF, and GIGA-WoLF algorithms for the coordination game. As expected, GIGA, PHC-WoLF, and GIGA-WoLF converges to one of the NE strategies faster, because WPL is slightly biased against pure strategies.

Figure 14 shows the convergence of previous GA-MARL algorithms for the matching pennies game. GIGA hopelessly oscillates around the NE, as expected. PHC-WoLF is better, but with relatively high standard deviation across runs (when compared to WPL). GIGA-WoLF has comparable performance to WPL, but GIGA-WoLF takes longer to converge (the width of oscillations around the NE is higher than WPL).

The tricky game deserves more attention as it is one of the challenging two-player-two-action games (Bowling & Veloso, 2002). Figure 15 shows the convergence of PHC-WoLF, GIGA, and GIGA-WoLF. Only GIGA-WoLF converges but with slower rate than our approach (Figure 12(c)). The performance of GIGA, PHC-WoLF, and GIGA-WoLF conforms to the results reported previously by their authors (Bowling & Veloso, 2002; Bowling, 2005). The remainder of our experiments (Section 5.3) focuses on GIGA-WoLF because it has performed the best among previous GA-MARL algorithms.

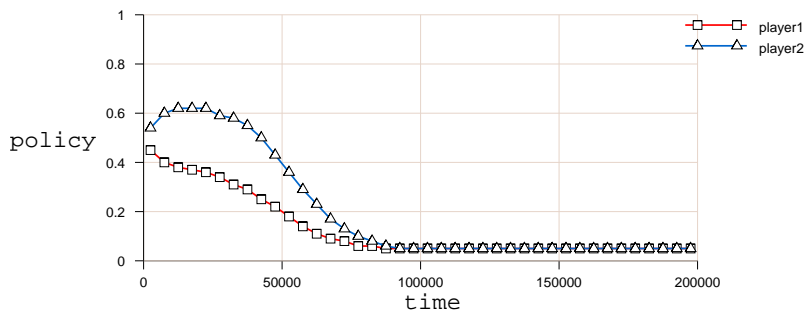

(a) WPL: coordination game

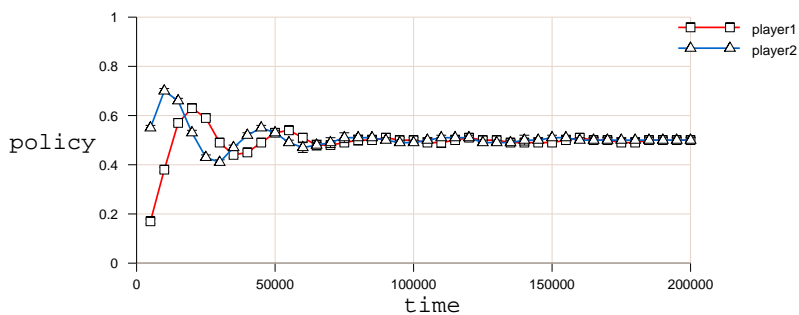

(b) WPL: matching pennies game

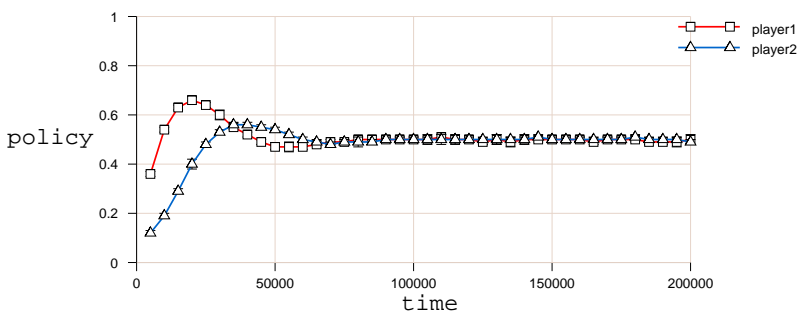

(c) WPL: tricky game

Figure 12: Convergence of WPL in benchmark two-player-two-action games. The figures plot the probability of playing the first action for each player (vertical axis) against time (horizontal axis).

## 5.3 Games Larger than 2x2

Figures 16 and 17 plot the policy of the row player over time for the games in Table 2 for both WPL and GIGA-WoLF when the initial joint strategy is $([0.1, 0.8, 0.1]_r, [0.8, 0.1, 0.1]_c)$ (we have also tried $([\frac{1}{3}, \frac{1}{3}, \frac{1}{3}]_r, [\frac{1}{3}, \frac{1}{3}, \frac{1}{3}]_c)$, which produced similar results), $\eta = 0.001$, and for two values of $\alpha$: 0.1 and 1. For the rock-paper-scissors game (Figure 16) both GIGA-WoLF and WPL converge when $\alpha = 0.1$, while only WPL converges when $\alpha = 1$. In Shapley's game (Figure 17) GIGA-WoLF keeps oscillating for both $\alpha = 1$ and $\alpha = 0.1$ (GIGA-WoLF's performance gets worse as $\alpha$ increases). WPL on the other hand performs better as $\alpha$ increases and converges in Shapley's game when $\alpha = 1$.

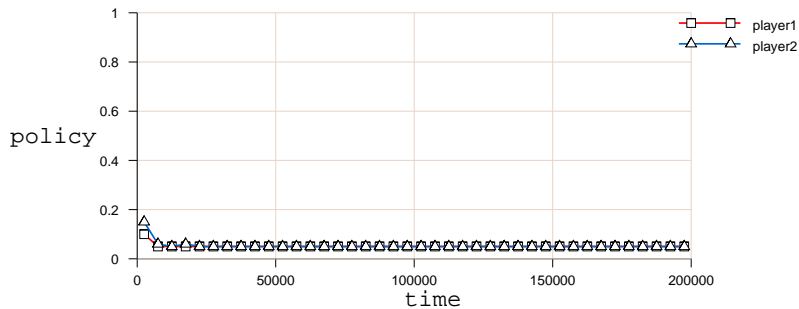

(a) PHC-WoLF: coordination game

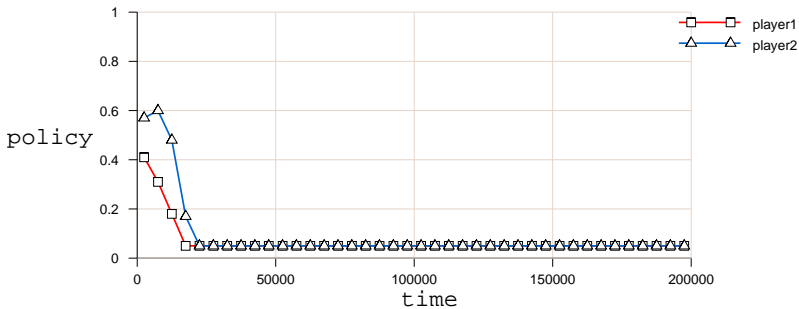

(b) GIGA: coordination game

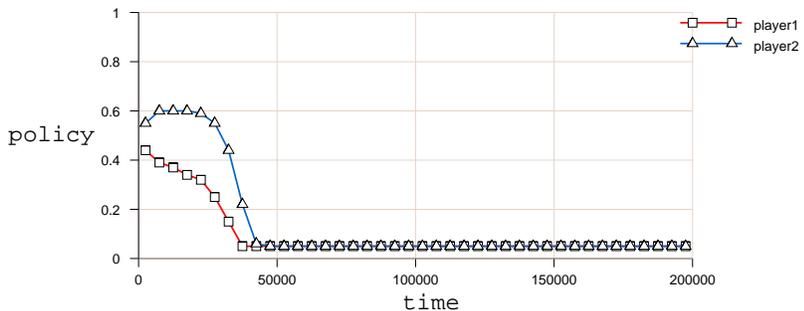

(c) GIGA-WoLF: coordination game

Figure 13: Convergence of the previous GA-MARL algorithms (GIGA, PHC-WoLF, and GIGA-WoLF) in the coordination game. The figures plot the probability of playing the first action for each player (vertical axis) against time (horizontal axis).

We believe the reason is that small $\alpha$ leads to an out-of-date reward estimate which in turn leads agents to continuously chase the NE without successfully converging. Smaller $\alpha$ means more samples contribute to the computed expected reward. Using more samples to estimate the reward makes the estimate more accurate. However, the time required to get more samples may in fact degrade the accuracy of reward estimate, because the expected reward changes over time.

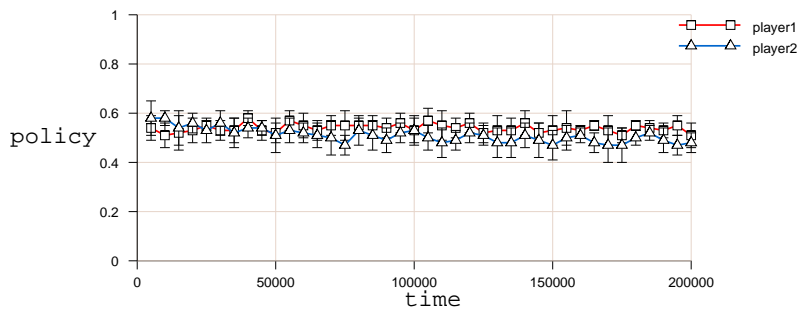

(a) PHC-WoLF: matching game

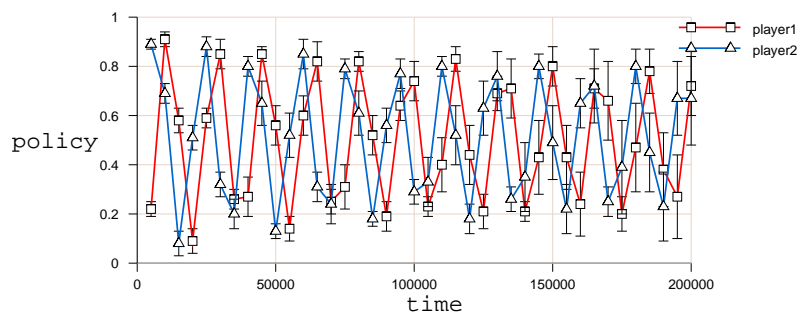

(b) GIGA: matching game

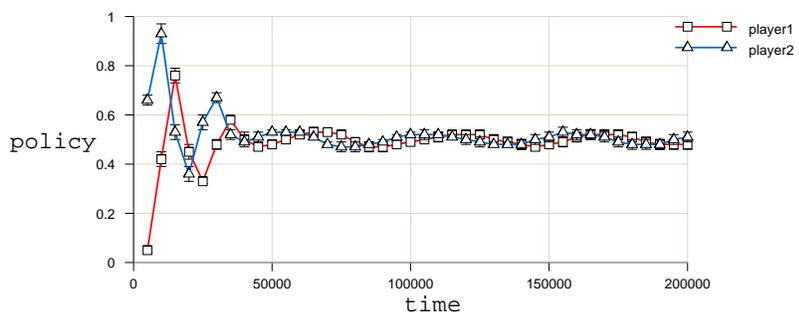

(c) GIGA-WoLF: matching game

Figure 14: Convergence of the GA-MARL algorithms (GIGA, PHC-WoLF, and GIGA-WoLF) in the match-
ing pennies game. The figures plot the probability of playing the first action for each player
(vertical axis) against time (horizontal axis).

Setting the value of $\alpha$ always to 1 can result in sub-optimal performance, as illustrated by the
following experiment. Consider the *biased* game shown in Table 3. The NE policy of the biased
game is mixed with probabilities not uniform across actions, unlike previous benchmark games that
had a mixed NE with uniform probability across actions. When $\alpha$ was set to 0.01, WPL converges
to a policy close to the NE policy as shown in Figure 18. When $\alpha = 1$, WPL converged to a policy
far from the NE policy as shown in Figure 19.

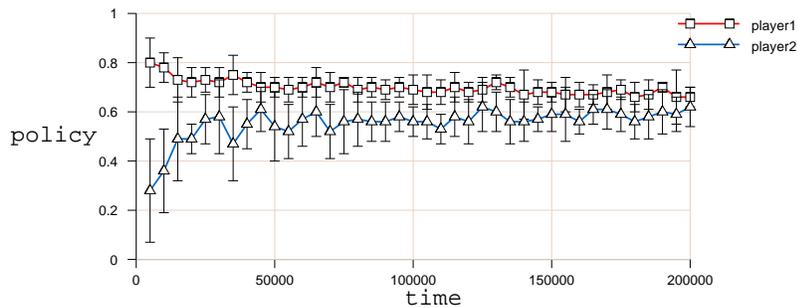

(a) PHC-WoLF: tricky game

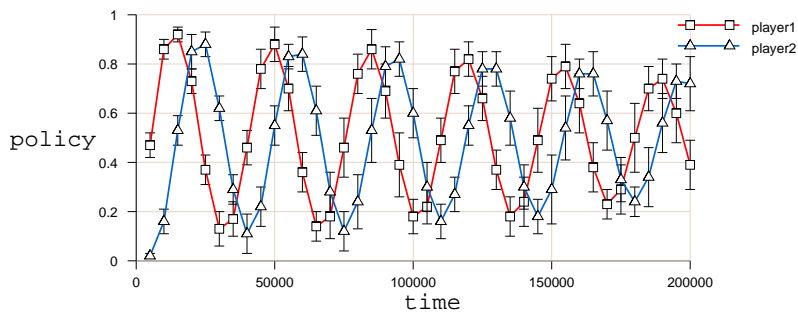

(b) GIGA: tricky game

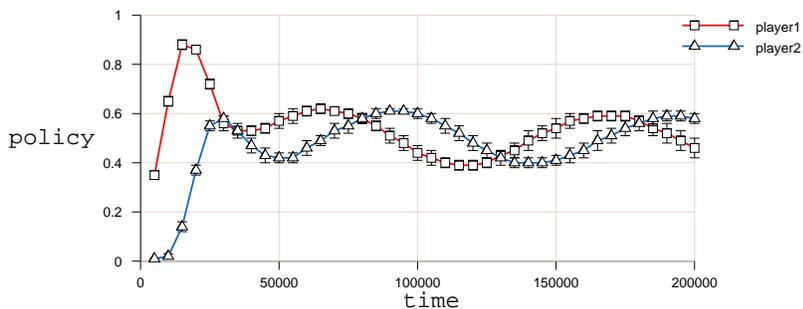

(c) GIGA-WoLF: tricky game

Figure 15: Convergence of the GA-MARL algorithms (GIGA, PHC-WoLF, and GIGA-WoLF) in the tricky game. The figures plot the probability of playing the first action for each player (vertical axis) against time (horizontal axis).

Table 3: Biased game: NE=(0.15,0.85) & (0.85,0.15)

|    | a1       | a2       |
|----|----------|----------|
| a1 | 1.0,1.85 | 1.85,1.0 |
| a2 | 1.15,1.0 | 1.00,1.15 |

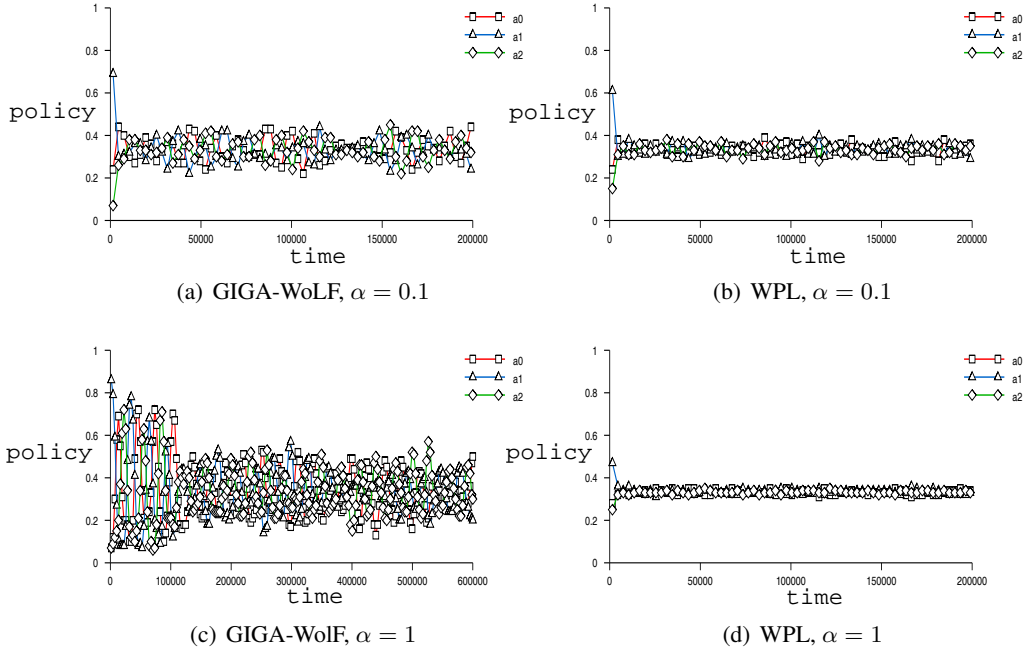

(a) GIGA-WoLF, $\alpha = 0.1$        (b) WPL, $\alpha = 0.1$

(c) GIGA-WoLF, $\alpha = 1$        (d) WPL, $\alpha = 1$

Figure 16: Convergence of GIGA-WoLF and WPL in the rock-paper-scissors game. The figures plot the probability of playing each action of the first player (vertical axis) against time (horizontal axis).

To understand the effect of having $\alpha = 1$ in case of WPL, consider the following scenario in the biased game. Suppose the policy of the column player is fixed at 0.7,0.3. If the value of $\alpha$ is small, then the action value function approximates well the expected value of each action. The value of the first row action (from the row player perspective) = $0.7 \times 1 + 0.3 \times 1.85 = 1.255$. Similarly, the value of the second row action = $0.7 \times 1.15 + 0.3 \times 1 = 1.105$. Unless the column player changes its policy, the row player gradient clearly points toward increasing the probability of choosing the first action (up to 1). Now consider the case when $\alpha = 1$. In that case the action value reflects the latest (sample) reward, not the average. In this case, the probability of choosing the first action, p, increases on average by

$$
\begin{aligned}
\overline{\Delta} = &\quad 0.7 \times [0.7 \times (1 - 1.15) \times p + 0.35 \times (1-1)] + 0.3 \times \\
&\quad [0.7 \times (1.85 - 1.15) \times (1 - p) + 0.3 \times (1.85 - 1) \times (1 - p)] \\
= &\quad -0.297p + 0.2235
\end{aligned}
$$

which means that the row player's policy will effectively stabilize when $\overline{\Delta} = 0$ or $p = 0.75$. In other words, it is possible for players to stabilize at an equilibrium that is not an NE. Note that this problem occurs mainly in biased games. In common benchmark games where the NE is uniform across actions or pure, WPL will still converge to an NE even if $\alpha = 1$, as shown in Figure 16 and Figure 17.

Tuning $\alpha$ is not an easy task when the expected reward itself is dynamically changing (because other agents are changing their policies concurrently). We are currently investigating an extension to WPL that automatically recognizes oscillations and adjusts the value of $\alpha$ accordingly. Preliminary results are promising.

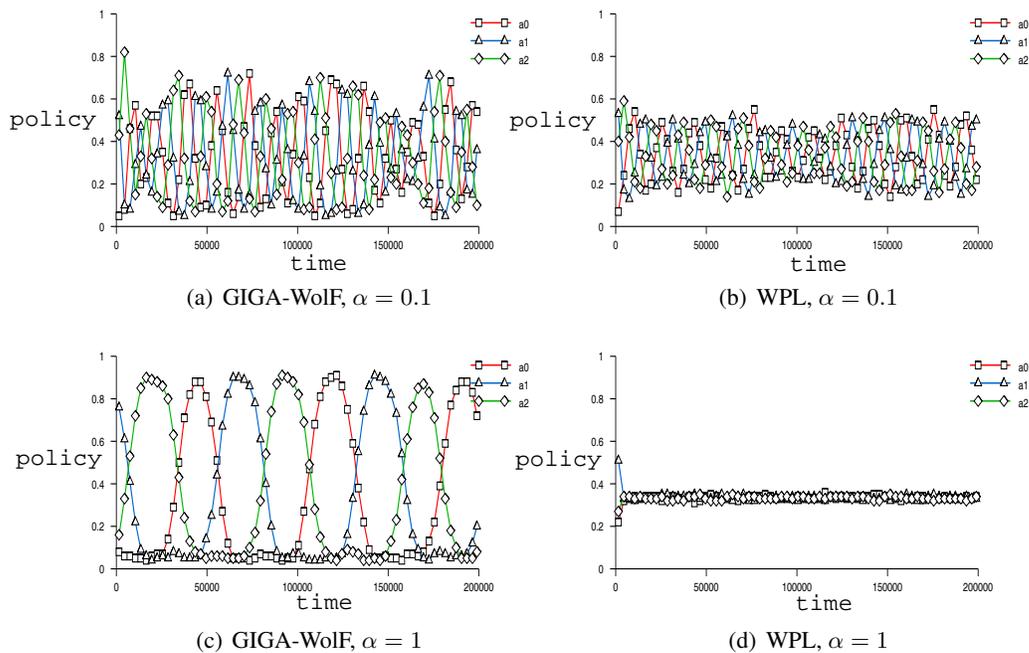

(a) GIGA-WoLF, $\alpha = 0.1$      (b) WPL, $\alpha = 0.1$

(c) GIGA-WoLF, $\alpha = 1$      (d) WPL, $\alpha = 1$

Figure 17: Convergence of GIGA-WoLF and WPL in Shapley's game. The figures plot the probability of playing each action of the first player (vertical axis) against time (horizontal axis).

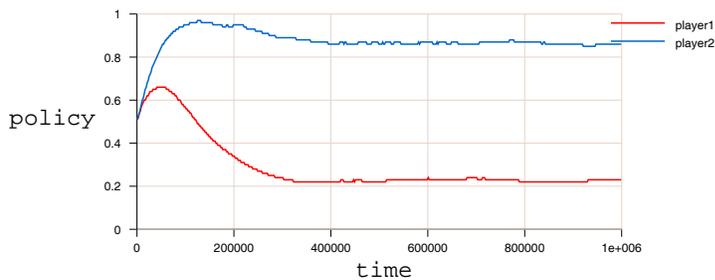

Figure 18: Convergence of WPL with fixed value-learning-rate of 0.01 for the biased game. Horizontal axis is time. Vertical axis is the probability of the first action for each player.

The following section illustrates the applicability of our algorithm by applying WPL in a more realistic setting involving 100 agents.

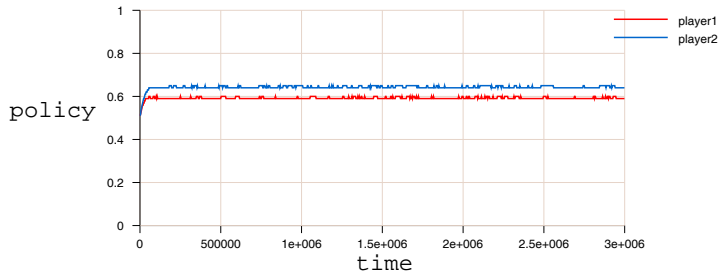

Figure 19: Convergence of WPL with fixed value-learning-rate of 1.0 for the biased game. Horizontal axis is time. Vertical axis is the probability of the first action for each player.

## 5.4 Distributed Task Allocation Problem (DTAP)

We use a simplified version of the distributed task allocation domain (DTAP) (Abdallah & Lesser, 2007), where the goal of the multiagent system is to assign tasks to agents such that the service time of each task is minimized. For illustration, consider the example scenario depicted in Figure 20. Agent A0 receives task T1, which can be executed by any of the agents A0, A1, A2, A3, and A4. All agents other than agent A4 are overloaded, and therefore the best option for agent A0 is to forward task T1 to agent A2 which in turn forwards the task to its left neighbor (A5) until task T1 reaches agent A4. Although agent A0 does not know that A4 is under-loaded (because agent A0 interacts only with its immediate neighbors), agent A0 will eventually learn (through experience and interaction with its neighbors) that sending task T1 to agent A2 is the best action without even knowing that agent A4 exists.

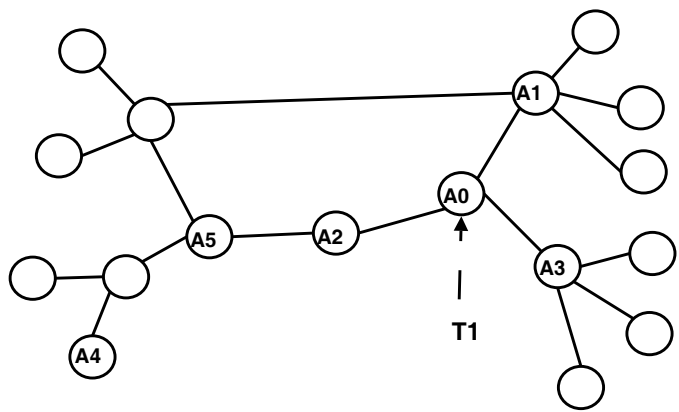

Figure 20: Task allocation using a network of agents.

Q-learning is not appropriate in such setting due to communication delay (which results in partial observability). For example, even if two neighbors have practically the same load, Q-learning will assign all incoming requests to one of the neighbors until a feedback is received later indicating change in the load. It should be noted that Q-learning was successfully used in the packet routing domain (Boyan & Littman, 1994; Dutta, Jennings, & Moreau, 2005), where load balancing is not the main concern (the main objective is routing a packet from a particular source to a particular destination).

Now let us define the different aspects of the DTAP domain more formally. Each time unit, agents make decisions regarding all task requests received during this time unit. For each task, the agent can either execute the task locally or send the task to a neighboring agent. If an agent decides to execute the task locally, the agent adds the task to its local queue, where tasks are executed on a first come first serve basis, with unlimited queue length.

Each agent has a physical location. Communication delay between two agents is proportional to the Euclidean distance between them, one time unit per distance unit. Agents interact via two types of messages. A REQUEST message $\langle i, j, T \rangle$ indicates a request sent from agent $i$ to agent $j$ requesting task $T$. An UPDATE message $\langle i, j, T, R \rangle$ indicates a feedback (reward signal) from agent $i$ to agent $j$ that task $T$ took $R$ time steps to complete (the time steps are computed since agent $i$ received $T$'s request).

The main goal of DTAP is to reduce the total service time, averaged over tasks, $ATST = \frac{\sum_{T \in \overline{T}_\tau} TST(T)}{|\overline{T}_\tau|}$, where $\overline{T}_\tau$ is the set of task requests received during a time period $\tau$ and TST is the total time a task spends in the system. TST consists of the time for routing a task request through the network, the time a task request spends in the local queue, and the time of actually executing the task.

It should be noted that although the underlying simulator have different underlying states, we deliberately made agent oblivious to these states. The only feedback an agent gets (consistent with our initial claim) is its own reward. The agents learn a joint policy that makes a good compromise over the different unobserved states (because the agents can not distinguish between these states).

We have evaluated WPL's performance using the following setting.[9] 100 agents are organized in a 10x10 grid. Communication delay between two adjacent agents is two time units. Tasks arrive at the 4x4 sub-grid at the center at rate 0.5 tasks/time unit. All agents can execute a task with a rate of 0.1 task/time unit (both task arrival and service durations follow an exponential distribution).

Figure 21 shows the results of applying GIGA, GIGA-WoLF and WPL using value-learning-rate of 1 and policy-learning-rate of 0.0001. GIGA fail to converge, while WPL and GIGA-WoLF do converge. WPL converges faster and to a better ATST: GIGA-WoLF's ATST converges to around 100 time units, while WPL's ATST converges to around 70 time units.

We believe GIGA-WoLF's slow convergence is due to the way GIGA-WoLF works. GIGA-WoLF relies on learning a slowly moving policy in addition to the actual policy $\pi$ in order to approximate the NE policy. This requires more time than the WPL algorithm. Furthermore, GIGA-WoLF's dynamics can be discontinuous prior to convergence and reach extreme deterministic policies even if the NE policy is mixed. In a large system, this can have ripple effect and slow system-wide convergence. WPL, on the other hand, has continuous dynamics, allowing faster collective convergence.

## 6. Conclusion and Future Work

This work presents WPL, a gradient ascent multiagent reinforcement learning algorithm (GA-MARL) that assumes an agent neither knows the underlying game nor observes other agents. We experimentally show that WPL converges in benchmark 2-player-2-action games. We also show that WPL converges in Shapley's game where none of the previous GA-MARL algorithms successfully converged. We verify the practicality of our algorithm in the distributed task allocation domain with a network of 100 agents. WPL outperforms the state-of-the-art GA-MARL algorithms in both the speed of convergence and the expected reward. We analyze the dynamics of WPL and show that it has continuous non-linear dynamics, while previous GA-MARL algorithms had discontinuous dynamics. We show that our predicted theoretical behavior is consistent with our experimental results.

---

9. The simulator is available online at `http://www.cs.umass.edu/~shario/dtap.html`.

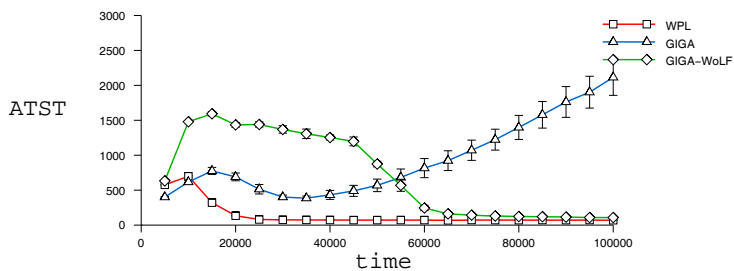

Figure 21: ATST of a 10x10 grid for different MARL algorithms. Horizontal axis is time while vertical axis is ATST.

In this work we briefly illustrated the importance of value-learning-rate and how it affects convergence, particularly for our proposed algorithm WPL. Finding the right balance and theoretically analyzing how it affects convergence are interesting research questions. We are currently investigating an extension to WPL that automatically finds a good value-learning-rate. Preliminary experimental results are promising.

Another future direction we are considering is extending our theoretical analysis to games with more actions and more players in order to verify our experimental findings in Shapley's game and the distributed task allocation domain. We are currently investigating alternative methodologies for analyzing dynamics, including the evolutionary dynamics (Tuyls, 't Hoen, & Vanschoenwinkel, 2006) and Lyapunov stability theory (Khalil, 2002).

## 7. Acknowledgments